\pdfoutput=1
\documentclass[11pt]{article}

\usepackage{authblk}
\usepackage{acl}

\usepackage{times}
\usepackage{latexsym}
\usepackage{multirow}
\usepackage{pgfplotstable}
\usepackage{multicol}
\usepackage{graphicx}
\usepackage{hyperref}
\usepackage{braket}
\usepackage{float}
\usepackage{amsmath}
\usepackage{etoolbox}

\usepackage{siunitx}
\usepackage[T1]{fontenc}

\usepackage[utf8]{inputenc}

\usepackage{microtype}

\newrobustcmd{\B}{\fontseries{b}\selectfont}
\newrobustcmd{\U}{\underline}

\newcommand\Tstrut{\rule{0pt}{3ex}}
\newcommand\Bstrut{\rule[-2ex]{0pt}{0pt}}

\title{Parameter-Efficient Legal Domain Adaptation}

\author[1]{Jonathan Li}
\author[1,2]{Rohan Bhambhoria}
\author[1,2]{Xiaodan Zhu}
\affil[1]{\hspace{1pt}Ingenuity Labs, Queen’s University}
\affil[2]{\hspace{1pt}Department of Electrical and Computer Engineering, Queen’s University}

\affil[ ]{\texttt{\{jxl, r.bhambhoria, xiaodan.zhu\}queensu.ca}}

\begin{document}
\maketitle
\begin{abstract}

Seeking legal advice is often expensive. Recent advancements in machine learning for solving complex problems can be leveraged to help make legal services more accessible to the public. However, real-life applications encounter significant challenges. State-of-the-art language models are growing increasingly large, making parameter-efficient learning increasingly important. Unfortunately, parameter-efficient methods perform poorly with small amounts of data \citep{gu-etal-2022-ppt}, which are common in the legal domain (where data labelling costs are high). To address these challenges, we propose parameter-efficient legal domain adaptation, which uses vast unsupervised legal data from public legal forums to perform legal pre-training. This method exceeds or matches the fewshot performance of existing models such as LEGAL-BERT \cite{chalkidis-etal-2020-legal} on various legal tasks while tuning only approximately 0.1\% of model parameters. Additionally, we show that our method can achieve calibration comparable to existing methods across several tasks. To the best of our knowledge, this work is among the first to explore parameter-efficient methods of tuning language models in the legal domain. 

\end{abstract}

\renewcommand{\floatpagefraction}{.8}%

\section{Introduction}

\begin{figure}
    \centering
    \includegraphics[scale=0.37]{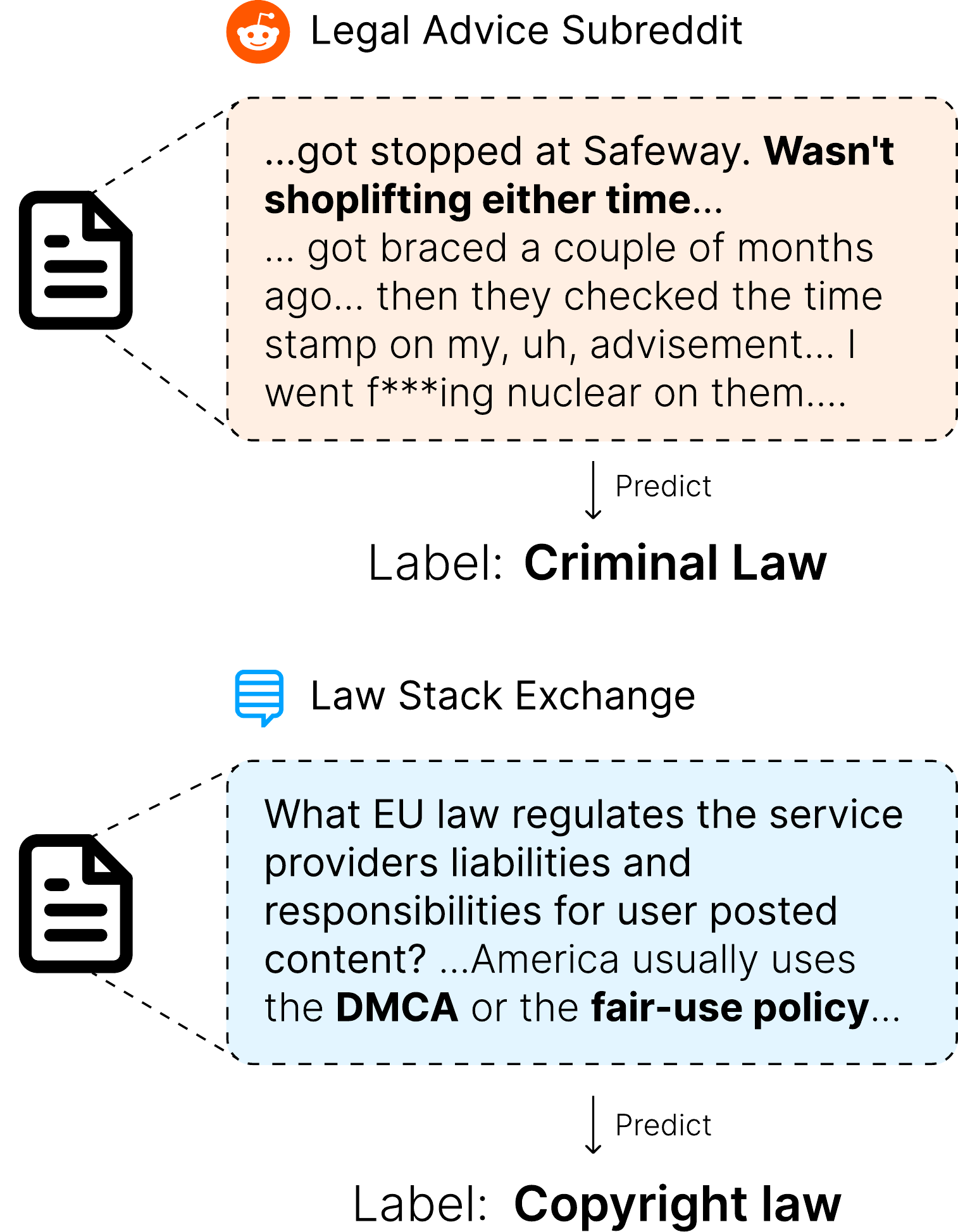}
    \caption{Example classification task using legal questions from Legal Advice Subreddit (top) and Law Stack Exchange (bottom). Reddit data is generally more informal than Stack Exchange.}
    \label{fig:taskExample}
\end{figure}

Seeking legal advice from lawyers can be expensive. However, a machine learning system that can help answer legal questions could greatly aid laypersons in making informed legal decisions. Existing legal forums, such as Legal Advice Reddit and Law Stack Exchange, are valuable data sources for various legal tasks. On one hand, they provide good sources of labelled data, such as mapping legal questions to their areas of law (for classification), as shown in Figure~\ref{fig:taskExample}. On the other hand, they contain hundreds of thousands of legal questions that can be leveraged for domain adaptation. Furthermore, questions on these forums can serve as a starting point for tasks that do not have labels found directly in the dataset, such as classifying the severity of a legal question. In this paper, we show that this vast unlabeled corpus can improve performance on question classification, opening up the possibility of studying other tasks on these public legal forums.

In the past few years, large language models have shown effectiveness in legal tasks \citep{chalkidis-etal-2022-lexglue}. A widespread method used to train these models is finetuning. Although finetuning is very effective, it is prohibitively expensive; training all the parameters requires large amounts of memory and requires a full copy of the language model to be saved for each task. Recently, prefix tuning (\citealp{li-liang-2021-prefix}; \citealp{liu-etal-2022-p}) has shown great promise by tuning under 1\% of the parameters and still achieving comparable performance to finetuning. Unfortunately, prefix tuning performs poorly in low-data (i.e., fewshot) settings \citep{gu-etal-2022-ppt}, which are common in the legal domain. Conveniently, domain adaptation using large public datasets is an ideal setting for the legal domain with abundant unlabelled data (from public forums) and limited labelled data. To this end, we introduce prefix domain adaptation, which performs domain adaptation for prompt tuning to improve fewshot performance on various legal tasks.

Overall, our main contributions are as follows:

\begin{itemize}
    \item We introduce prefix adaptation, a method of domain adaptation using a prompt-based learning approach.
    
    \item We show empirically that performance and calibration of prefix adaptation matches or exceeds LEGAL-BERT in fewshot settings while only tuning approximately 0.1\% of the model parameters.
    
    \item We contribute two new datasets to facilitate different legal NLP tasks on the questions asked by laypersons, towards the ultimate objective of helping make legal services more accessible to the public.
\end{itemize}

\section{Related Works}

\paragraph{Forums-based Datasets} Public forums have been used extensively as sources of data for machine learning. Sites like Stack Overflow and Quora have been used for duplicate question detection (\citealp{wang-etal-2022-duplicate}; \citealp{sharma-etal-2019-natural}). Additionally, many prior works have used posts from specific sub-communities (called a "subreddit") on Reddit for NLP tasks, likely due to the diversity of communities and large amount of data provided. \citet{barnes-etal-2021-dank} used a large number of internet memes from multiple meme-related subreddits to predict how likely a meme is to be popular. Other works, such as \citet{basaldella-etal-2020-cometa}, label posts from biomedical subreddits for biomedical entity linking. Similar to the legal judgement prediction task, \citet{lourie-etal-2021-scruples} suggest using "crowdsourced data" from Reddit to perform ethical judgement prediction; that is, they use votes from the "r/AmITheAsshole" subreddit to classify who is "in the wrong" for a given real-life anecdote. We explore using data from Stack Exchange and Reddit, which has been vastly underexplored in previous works for the legal domain.

\paragraph{Full Domain Adaptation} Previous works such as BioBERT \citep{lee-etal-2019-biobert} and SciBERT \citep{beltagy-etal-2019-scibert} have shown positive results while domain adapting models. In the industry, companies often use full domain adaptation for legal applications \footnote{\href{https://vectorinstitute.ai/2020/04/02/how-thomson-reuters-uses-nlp-to-enable-knowledge-workers-to-make-faster-and-more-accurate-business-decisions/}{https://vectorinstitute.ai/2020/04/02/how-thomson-reuters-uses-nlp-to-enable-knowledge-workers-to-make-faster-and-more-accurate-business-decisions/}}. \citet{chalkidis-etal-2020-legal} introduce LEGAL-BERT, a BERT-like model domain adapted for legal tasks. They show improvements across various legal tasks by training on a domain-specific corpus. \citet{zheng-etal-2021-when} also perform legal domain adapation, using the Harvard Law case corpus, showing better performance in the CaseHOLD multiple-choice question answering task. Unlike existing works, we perform domain adaptation parameter-efficiently, showing similar performance in a fewshot setting. We compare our approach against LEGAL-BERT as a strong baseline.

\paragraph{Parameter-efficient Learning} Language models have scaled to over billions of parameters (\citealp{he2021deberta}; \citealp{NEURIPS2020_1457c0d6}), making research memory and storage intensive. Recently, parameter-efficient training methods---techniques that focus on tuning a small percentage of the parameters in a neural network---have been a prominent research topic in natural language processing. More recently, prefix tuning \citep{li-liang-2021-prefix} has attracted much attention due to its simplicity, ease of implementation, and effectiveness. In this paper, we use P-Tuning v2 \citep{liu-etal-2022-p}, which includes an implementation of prefix tuning.

Previously, \citet{gu-etal-2022-ppt} explored improving prefix tuning's fewshot performance with pre-training by rewriting downstream tasks for a multiple choice answering task (in their "unified PPT"), and synthesizing multiple choice pre-training data (from OpenWebText). Unlike them, we focus on domain adaptation and not general pre-training. We show a much simpler method of prompt pre-training using the masked language modelling (MLM) task while preserving the format of downstream tasks. \citet{ge-etal-2022-domain} domain adapt continuous prompts (not prefix tuning) to improve performance with vision-transformer models for different image types (e.g., "clipart", "photo", or "product").

\citet{zhang-etal-2021-unsupervised-domain} domain adapt an adapter \citep{houlsby-etal-2019-parameter}, which is another type of parameter-efficient training method where small neural networks put between layers of the large language model are trained.  \citet{vu-etal-2022-spot} explored the transferability of prompts between tasks. They trained a general prompt for the "prefix LM" \citep{JMLR:v21:20-074} objective on the Colossal Clean Crawled Corpus \citep{JMLR:v21:20-074}. They do not study the efficacy of their general-purpose prompt in fewshot scenarios. Though we use a similar unsupervised language modelling task \citep{devlin-etal-2019-bert}, we aim to train a domain adapted prompt and not a general-purpose prompt.

\begin{figure*}[ht]
    \centering
    \includegraphics[width=\textwidth]{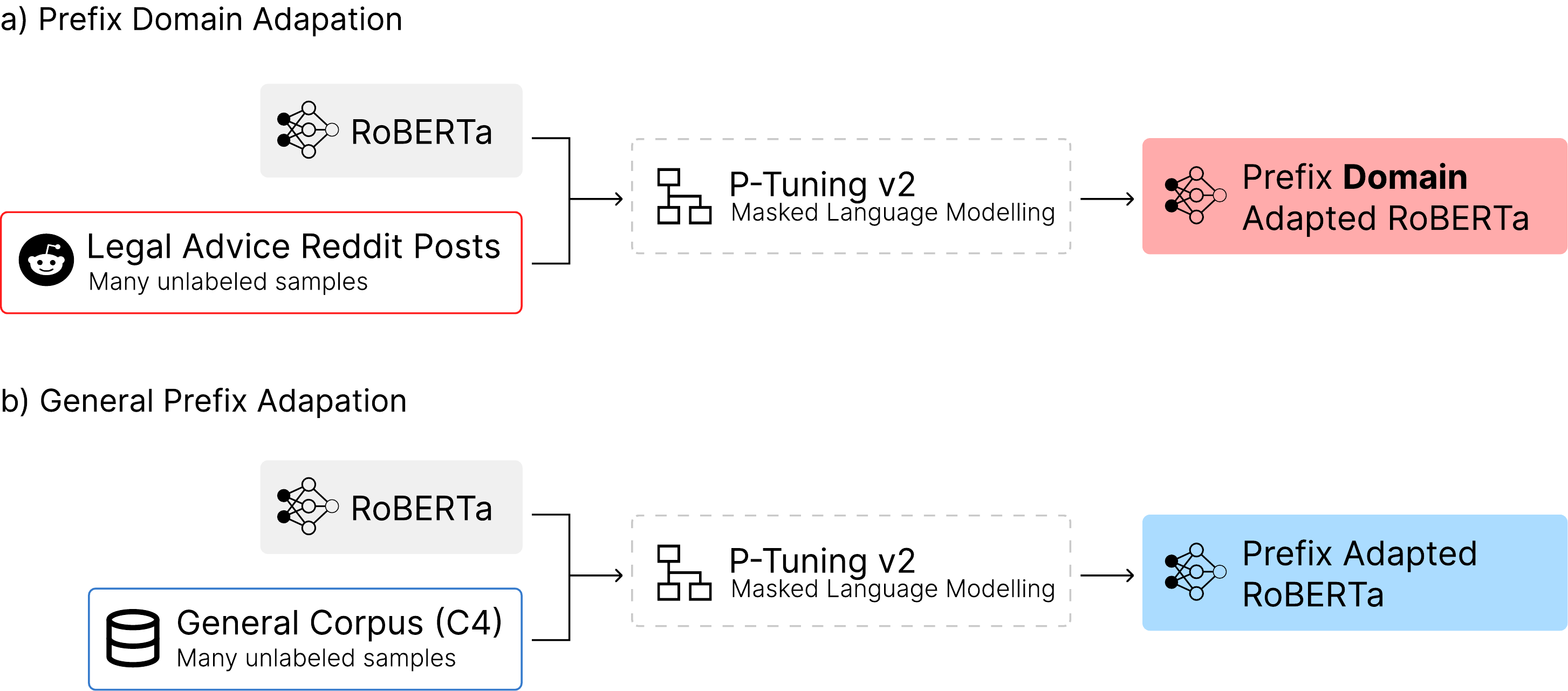}
    \caption{Training process for our methods, with colored boxes representing model weights, colored outlines representing datasets, and dotted outlines representing training method (in this case, P-Tuning v2). Notice that (a) prefix domain adapation and (b) prefix adaptation both use the same starting model and training method, but different datasets.}
    \label{fig:trainingProcess}
\end{figure*}

\section{Background}

\paragraph{Legal Forums} Seeking legal advice from a lawyer can be incredibly expensive. However, public legal forums are incredibly accessible to laypersons to ask legal questions. One popular community is the \href{https://www.reddit.com/r/legaladvice/}{Legal Advice Reddit} community (2M+ members), where users can freely ask personal legal questions. Typically, the questions asked on the Legal Advice Subreddit are written informally and receive informal answers. Another forum is the \href{https://law.stackexchange.com/}{Law Stack Exchange}, a community for questions about the law. Questions are more formal than on Reddit. Additionally, users are not allowed to ask about a specific case and must ask about law more hypothetically, as specified in the rules.

In particular, data from the Legal Advice Subreddit is especially helpful in training machine learning models to help laypersons in law, as questions are in the format and language that regular people would write in (see Figure~\ref{fig:taskExample}). We run experiments on Law Stack Exchange (LSE) for comprehensiveness, though we believe that the non-personal nature of LSE data makes it less valuable than Reddit data in helping laypersons.

\paragraph{Prefix Tuning} As language models grow very large, storage and memory constraints make training impractical or very expensive. Deep prefix tuning addresses these issues by prepending continuous prompts to the transformer. These continuous prefix prompts, which are prepended to each attention layer in the model, and a task-specific linear head (such as a classification head) are trained.

More formally, for each attention layer $L_{i}$ (as per \citealp{vaswani-etal-2017-attention}) in BERT's encoder, we append some trainable prefixes $P_{k}$ (trained key prefix) and $P_{v}$ (trained value prefix) with length $n$ to the key and value matrices for some initial prompts:

\begin{flalign}
\begin{split}
    L_{i}=\mathbf{Attn}(&xW_{q}^{(i)},\\
    &Cat(P_{k}^{(i)}, xW_{k}^{(i)}),\\
    &Cat(P_{v}^{(i)}, xW_{v}^{(i)}))
\end{split}
\label{eqn:prefixTuning}
\end{flalign}

With $W_{\Set{q, k, v}}^{(i)}$ representing the respective query, key, or value matrices for the attention at layer $i$, and $x$ denoting the input to layer $i$. Here, we assume single-headed attention for simplicity. Here, the $Cat$ function concatenates the two matrices along the dimension corresponding to the sequence length.

Note that in Equation~\ref{eqn:prefixTuning} we do not need to left-pad any query values, as the shape of the query matrix does not need to match that of the key and value matrices.

\paragraph{Expected Calibration Error} First suggested in \citet{naeini-etal-2015-obtaining} and later used for neural networks in \citet{guo-etal-2017-on}, expected calibration error (ECE) can determine how well a model is calibrated. In other words, ECE evaluates how closely a model's logit weights reflect the actual accuracy for that prediction. Calibration is important for two main reasons. First, having a properly calibrated model reduces misuse of the model; if output logits accurately reflect their real-world likelihood, then software systems using such models can better handle cases where the model is uncertain. Second, better calibration improves the interpretability of a model as we can better understand how confident a model is under different scenarios \citep{guo-etal-2017-on}. \citet{bhambhoria-etal-2022-interpretable} used ECE in the legal domain, where it is especially important due the high-stakes nature of legal decision making.

\section{Methods}

Here we outline our approach and other baselines for comparison.

\paragraph{RoBERTA} To establish a baseline, we train RoBERTa \citep{liu-etal-2019-roberta} for downstream tasks using full model tuning (referred to as "full finetuning"). In addition to the state of the art performance that RoBERTa achieves in many general NLP tasks, it has also shown very strong performance in legal tasks (\citealp{shaheen-etal-2020-large}; \citealp{bhambhoria-etal-2022-interpretable}).
Unlike some transformer models, RoBERTa has an encoder-only architecture, and is normally pre-trained on the masked language modelling task \citep{devlin-etal-2019-bert}. We evaluate the model on both of its size variants, RoBERTa-base (approximately 125M parameters) and RoBERTa-large (approximately 335M parameters).

\paragraph{LEGAL-BERT} We evaluate the effectiveness of our approach against LEGAL-BERT, a fully domain-adapted version of BERT for the legal domain \citep{chalkidis-etal-2020-legal}. In our experiments, we further perform full finetuning for each downstream task. The number of parameters in LEGAL-BERT (109M) is comparable to RoBERTa-base (125M), as used in our other experiments.

\paragraph{Full Domain Adaptation} We also perform full domain adaptation by pre-training all model parameters using the masked language modelling (MLM) task with text from each dataset. Then, we train this fully domain adapted model using full-model tuning for each downstream task. This method is a strong baseline for comparison, as we tune all model parameters twice (MLM pre-training and downstream task) for each task, taking up many computational resources.

\paragraph{P-Tuning v2} We compare our approach against P-Tuning v2 \citep{liu-etal-2022-p}, an "alternative to finetuning" that only optimizes a fractional percentage of parameters (0.1\%-3\%). It works by freezing the entire model, then appending some frozen prompts in each layer. That is, trainable prompts are added as prefixes to each layer, with only the key and value matrices of the self-attention mechanism trained. We use P-Tuning v2 as a baseline, being the original parameter-efficient training method that we base our study on.

\begin{figure*}[ht]
    \centering
    \includegraphics[width=\textwidth]{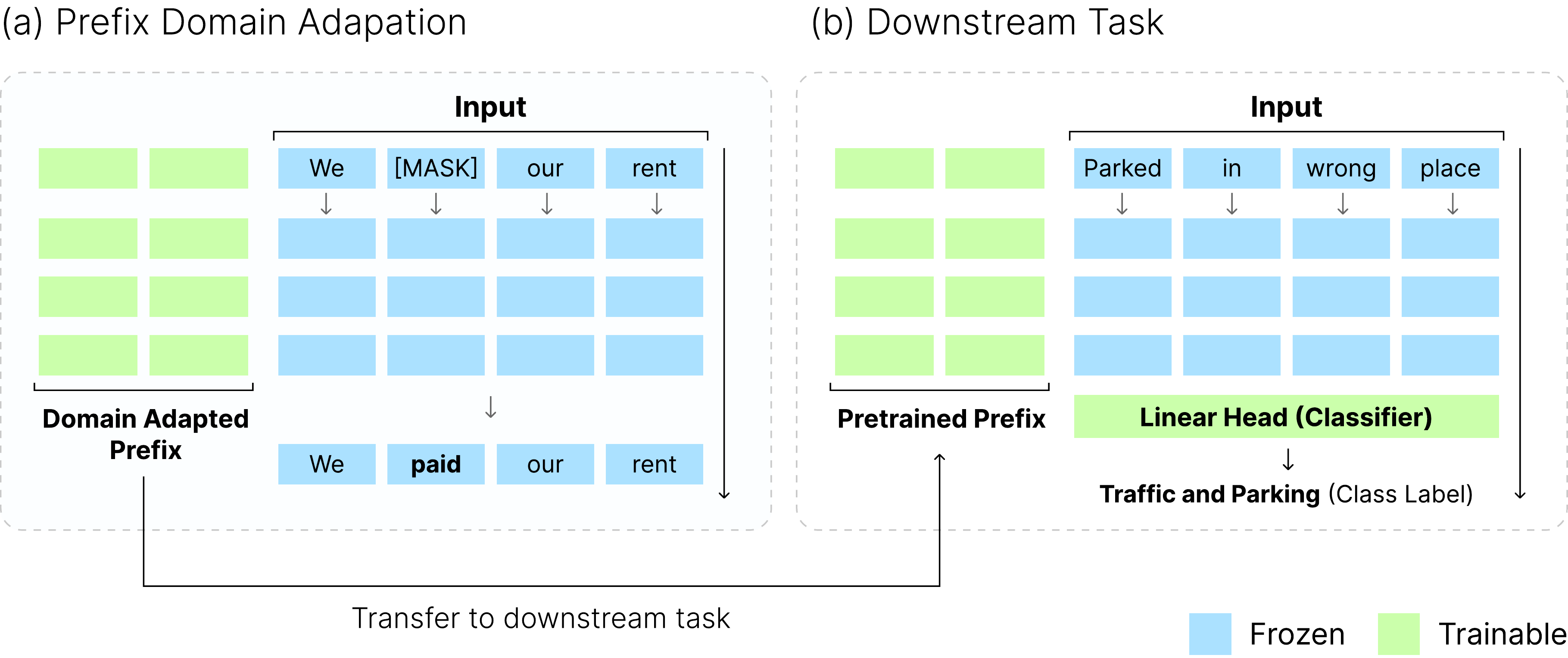}
    \caption{Toy example of how our framework works. (a) We pre-train a prefix on the unsupervised masked language modelling task. (b) We randomly initialize the classification head but preserve the prefix for downstream tasks. Green blocks represent trainable prompt embeddings or layers of the transformer, and blue blocks represent frozen embeddings or computations.}
    \label{fig:transformerLevelDiagram}
\end{figure*}

\paragraph{Prefix Domain Adaptation} Inspired by domain adaptation, we introduce \textit{prefix domain adaptation}, which domain adapts a deep prompt \citep{li-liang-2021-prefix} to better initialize it for downstream tasks. As the domain adapted deep prompt is very small (approximately 0.1\% the size of the base model), it is easy to store and distribute. Once trained, the deep prompt is used as a starting point for downstream tasks. 

More specifically, we train a deep prompt, using prefix tuning as in \citet{liu-etal-2022-p}\footnote{Same implementation as provided in \citet{liu-etal-2022-p}}, for the masked language modelling task \citep{devlin-etal-2019-bert} on a large, domain-specific unsupervised corpus, as shown in Figure~\ref{fig:trainingProcess}(a). Next, we use this pre-trained prompt and randomly initialize a task-specific head (such as a classification head for a classification task) for each downstream task. Finally, we train the resulting model for the downstream task, using the same prompt tuning approach from \citet{liu-etal-2022-p}. To the best of our knowledge, no prior works have trained a prefix prompt for a specific domain to better initialize it for downstream tasks using an unsupervised pre-training task (masked language modelling).

Formally, we can treat a prefix-tuned model as having a trained prefix $P$, and a trained task-specific head $H$. We group each downstream task into $m$ domains in $\Set{\mathcal{D}_1, \mathcal{D}_2, ..., \mathcal{D}_m}$, such that there is some overlap between the tasks in each domain $\mathcal{D}_i$. For each domain $\mathcal{D}_i$, we use a domain-specific corpus, $C_i$, to train some prefix $P_i$ for the masked language modelling task with prompt tuning (Figure~\ref{fig:transformerLevelDiagram}(a)). Then, for each downstream task in $\mathcal{D}_i$, we use the deep prefix $P_i$ to initialize the prompts, while randomly initializing the task-specific head $H_i$ (Figure~\ref{fig:transformerLevelDiagram}(b)).

\paragraph{Prefix Adaptation} In addition to prefix domain adaptation, we conduct experiments using our approach in general settings, inspired by work done in \citet{vu-etal-2022-spot} and \citet{gu-etal-2022-ppt}. We name this more general approach \textit{prefix adaptation}. That is, we test the performance of initializing a prompt with the masked language modelling task on a subset of the Colossal Clean Crawled Corpus \citep{JMLR:v21:20-074}, instead of domain-specific texts (illustrated in Figure~\ref{fig:trainingProcess}(b)). Formally, we use the same prefix domain adaptation approach as previously mentioned, but we group all tasks under one "General" domain $\mathcal{D}$, and thus only train one prefix $P$.

\section{Datasets}

We evaluate each of the approaches listed above on three different datasets.

\paragraph{Legal Advice Reddit} We introduce a new dataset from the Legal Advice Reddit community (known as "/r/legaldvice"), sourcing the Reddit posts from the Pushshift Reddit dataset \citep{baumgartner-etal-2020-the} \footnote{\href{https://huggingface.co/datasets/jonathanli/legal-advice-reddit}{https://huggingface.co/datasets/jonathanli/legal-advice-reddit}}. The dataset maps the text and title of each legal question posted into one of eleven classes, based on the original Reddit post's "flair" (i.e., tag). Questions are typically informal and use non-legal-specific language. Per the Legal Advice Reddit rules, posts must be about actual personal circumstances or situations. We limit the number of labels to the top eleven classes and remove the other samples from the dataset (more details in Appendix~\ref{sec:dataDetails}). To prefix adapt the model for Reddit posts, we use samples from the Legal Advice sub-reddit that are not labelled or do not fall under the top eleven classes. We use the provided "flair" for each question for a legal area classification task \citep{soh-etal-2019-legal}, as illustrated in Figure~\ref{fig:taskExample}.

\paragraph{European Court of Human Rights} We use the European Court of Human Rights (ECHR) dataset \citep{chalkidis-etal-2019-neural}, which consists of a list of facts specific to a legal case, labelled with violated human rights articles (if any). Specifically, we evaluate our approach on the binary violation prediction task, where the task is to predict whether a given case violates any human rights articles given a list of facts. We undersample this relatively large dataset to simulate a fewshot learning environment. To prefix adapt the model for ECHR cases, we use the original corpus of unlabelled cases (similar to what was done in \citealp{chalkidis-etal-2020-legal}). As the average document length is 700 words (above BERT's maximum length limit), we truncate the text to 500 tokens, concatenating the title and facts of the case together.

\paragraph{Law Stack Exchange} We also introduce a second dataset with data from the Law Stack Exchange (LSE)\footnote{\href{https://huggingface.co/datasets/jonathanli/law-stack-exchange}{https://huggingface.co/datasets/jonathanli/law-stack-exchange}}. This dataset is composed of questions from the \href{https://law.stackexchange.com/}{Law Stack Exchange}, which is a community forum-based website containing questions with answers to legal questions. Unlike the Legal Advice Reddit dataset, the Law Stack Exchange dataset is generally more formal (shown in Figure~\ref{fig:taskExample}), and questions are generally more theoretical or hypothetical. We link the questions with their associated tags (e.g., "copyright" or "criminal-law"), and perform the multi-label classification task. Though posts can have multiple tags, we use the questions with only one tag in the top 16 most frequent tags (excluding tags associated with countries). Similarly to the Legal Advice Reddit dataset, we use other unused questions from the Law Stack Exchange to prefix domain adapt the model.

\section{Experimental Setup}
We test our approaches under a fewshot setting, where prompt tuning is known to perform poorly \citep{gu-etal-2022-ppt}. We use RoBERTa-base and RoBERTa-large \citep{liu-etal-2019-roberta} for our experiments. To simulate a fewshot learning scenario, we randomly undersample the train and validation sets for each dataset, ensuring that the distribution of train and validation data roughly matches. Additionally, we vary the amount of data undersampled to study how fewshot size affects performance. In these tasks, we use a validation size of 256 (much smaller than the original) to represent true fewshot learning better \citep{perez-etal-2021-true}. Considering that fewshot learning is quite unstable, we ran all of our experiments five times, using the seeds $\set{10, 20, 30, 40, 50}$. We provide more training details in Appendix~\ref{sec:trainingDetails}.

\begin{table}
\centering
\resizebox{\linewidth}{!}{\begin{tabular}{c|c|c}
\hline
Dataset Name & $N_{class}$ & Fewshot Sizes \Tstrut\Bstrut\\
\hline
ECHR & 2 & 4, 8, 16, 32 \Tstrut\\
Legal Advice Reddit & 11 & 32, 64, 128, 256 \\
Law Stack Exchange & 16 & 32, 64, 128, 256 \Bstrut \\
\hline
\end{tabular}}
\caption{Classification tasks evaluated in our experiments. $N_{class}$ represents the number of classes, and "Fewshot Sizes" represents the various number of samples used (4 different fewshot sizes evaluated for each dataset).}
\label{tab:tasks}
\end{table}

There is often confusion around whether fewshot sizes represent the number of samples per class or the total number of samples \citep{perez-etal-2021-true}. In our results, the fewshot sizes we show are the exact number of training samples used (i.e., total training samples). The exact number of samples is listed in Table~\ref{tab:tasks}. To keep the number of samples per class roughly equivalent, we use fewer total samples for the ECHR task, which only has two classes.

\section{Results and Discussion}

\begin{table*}
\centering
\resizebox{\textwidth}{!}{\begin{tabular}{c|cccc|cccc|cccc}
\hline
& \multicolumn{4}{c|}{\textbf{Legal Advice Reddit}} & \multicolumn{4}{c|}{\textbf{Law Stack Exchange}} & \multicolumn{4}{c}{\textbf{European Court of Human Rights}} \Tstrut\\[2pt]
Fewshot Size & \textit{32} & \textit{64} & \textit{128} & \textit{256} & \textit{32} & \textit{64} & \textit{128} & \textit{256} & \textit{4} & \textit{8} & \textit{16} & \textit{32} \Bstrut\\
\hline
FT & \textbf{44.8\textsubscript{1.9}} & 56.7\textsubscript{9.4} & 63.8\textsubscript{2.8} & \underline{72.7\textsubscript{1.8}} & 19.5\textsubscript{17.1} & 29.0\textsubscript{14.8} & \underline{58.8\textsubscript{0.8}} & \underline{67.4\textsubscript{0.9}} & 53.7\textsubscript{1.6} & 60.1\textsubscript{5.5} & 66.5\textsubscript{8.7} & 66.3\textsubscript{3.5} \Tstrut\\
LEGAL-BERT + FT & 36.1\textsubscript{2.9} & 35.2\textsubscript{16.1} & 49.5\textsubscript{3.7} & 70.2\textsubscript{1.7} & 24.6\textsubscript{13.1} & 51.2\textsubscript{0.9} & 47.6\textsubscript{24.9} & \textbf{67.5\textsubscript{0.2}} & 59.3\textsubscript{12.4} & 55.8\textsubscript{3.8} & 61.1\textsubscript{8.7} & \underline{67.6\textsubscript{3.6}} \\
Domain Adapt + FT & 31.8\textsubscript{16.4} & \textbf{66.7\textsubscript{3.3}} & \underline{66.6\textsubscript{1.5}} & \textbf{75.8\textsubscript{0.9}} & \textbf{38.5\textsubscript{0.4}} & \textbf{53.2\textsubscript{2.5}} & \textbf{62.4\textsubscript{1.1}} & 66.6\textsubscript{0.6} & 47.6\textsubscript{2.3} & 51.2\textsubscript{1.7} & 47.9\textsubscript{1.3} & 56.7\textsubscript{2.2} \Bstrut\\
\hline
Prefix Domain Adapt & \underline{41.9\textsubscript{2.2}} & \underline{61.7\textsubscript{2.7}} & \textbf{66.6\textsubscript{0.6}} & 72.0\textsubscript{1.9} & \underline{36.1\textsubscript{0.9}} & \underline{52.4\textsubscript{1.7}} & 56.1\textsubscript{0.8} & 63.1\textsubscript{1.7} & \textbf{72.7\textsubscript{4.6}} & \underline{70.9\textsubscript{2.3}} & \textbf{75.1\textsubscript{1.8}} & \textbf{69.4\textsubscript{2.0}} \Tstrut\\
Prefix Adapt & 35.5\textsubscript{2.1} & 58.0\textsubscript{6.4} & 52.7\textsubscript{21.4} & 72.2\textsubscript{0.5} & 31.7\textsubscript{1.2} & 46.8\textsubscript{2.5} & 57.0\textsubscript{0.8} & 66.6\textsubscript{0.5} & 68.9\textsubscript{6.4} & \textbf{71.4\textsubscript{1.4}} & \underline{75.0\textsubscript{2.7}} & 66.3\textsubscript{7.8} \\
P-Tuning v2& 25.6\textsubscript{1.0} & 41.0\textsubscript{2.1} & 62.0\textsubscript{1.6} & 71.2\textsubscript{0.6} & 24.6\textsubscript{2.2} & 45.3\textsubscript{2.0} & 56.3\textsubscript{0.7} & 65.3\textsubscript{0.6} & \underline{70.9\textsubscript{2.6}} & 70.5\textsubscript{3.6} & 70.9\textsubscript{2.3} & 67.1\textsubscript{0.9} \Bstrut \\
\hline
\end{tabular}}
\caption{Classification results with RoBERTa-base (or similarly sized models), with fewshot size listed as italic numbers in the second row. Experiments run five times with different seeds, with subscripts representing the standard deviation of the five runs. \textbf{Bolded} results represent the best performance for the fewshot size, and \underline{underlined} results represent second best. All methods are assumed to be initialized from RoBERTa-base, except for LEGAL-BERT from \citet{chalkidis-etal-2020-legal}. "FT" represents fully finetuned for downstream tasks and "Domain Adapt" is full domain adaptation, with a line separating full-model (top) and parameter efficient (bottom) tuning methods.}
\label{tab:classification-pt}
\end{table*}

\begin{figure}
    \centering
    \includegraphics[width=\linewidth]{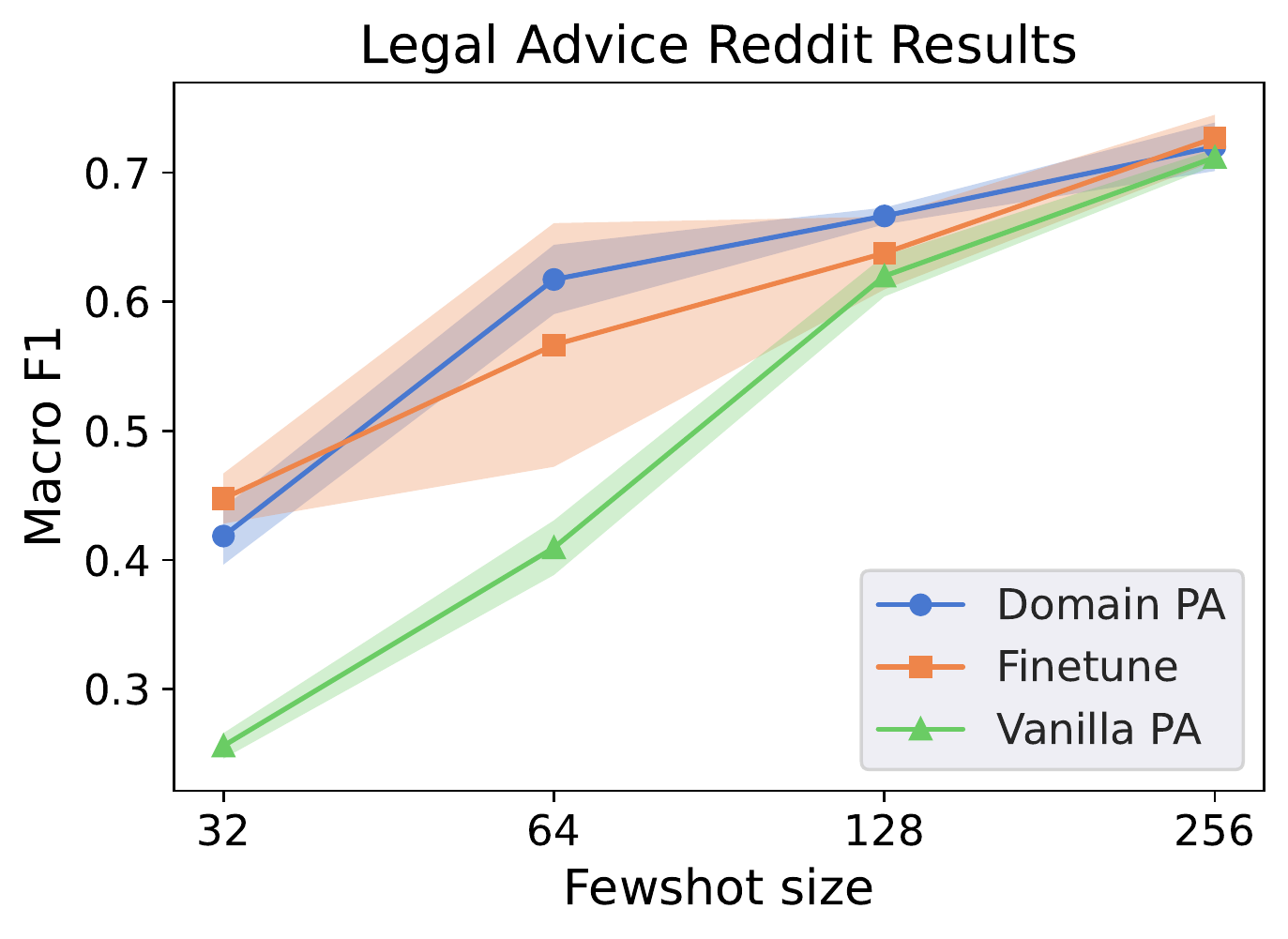}
    \caption{Various fewshot sizes and their performance (measured by macro F1). The shaded region represents the standard deviation across runs, while each point represents the mean performance across runs. Overall, our approach (prefix domain adaptation) matches the performance of full finetuning.}
    \label{fig:varyFewshotSizeResults}
\end{figure}

We make a few observations on our results, shown in Table~\ref{tab:classification-pt}. We observe that our method, prefix domain adaptation, outperforms both regular prefix tuning and full finetuning in most tasks across fewshot sizes, despite training considerably fewer parameters. We find that prefix adaptation is comparable to full domain adaptation; in some settings (such as ECHR and some Reddit fewshot settings), prefix adaptation even outperforms full domain adaptation. We argue that prefix domain adaptation achieves better fewshot performance relative to regular prefix tuning because the pre-trained prompts are closer to an effective prompt after our domain adaptation step. This is similar to full domain adaptation, which improves performance on downstream tasks relative to a base model \citep{chalkidis-etal-2020-legal} by making parameters closer to optimal parameters. Consistent with \citet{gu-etal-2022-ppt}, we find that regular prefix tuning falls behind full parameter tuning in fewshot settings.

Additionally, we find that LEGAL-BERT performs worse than other techniques on datasets with more informal language (such as the Reddit dataset). LEGAL-BERT shows more instability across seeds (i.e., larger standard deviation) . As LEGAL-BERT-SC (the model we use) was only trained on very formal legal text, it did not see many colloquialisms or slang during training that are prevalent in informal text. For this reason, we do not think LEGAL-BERT would be effective as initialization for tasks involving legal questions asked by laypersons, which typically do not use incredibly formal legal language.

In contrast to other datasets, the ECHR dataset's train and test split have different distributions. In fewshot scenarios with very little data (i.e., 4-16 examples), we find that prefix tuning based approaches perform better than full finetuning; this suggests that prefix tuning approaches are more robust to changes in distribution (and possibly noise). We also note that BERT with truncation (maximum token length of 500) performs a lot better than initially reported in \citet{chalkidis-etal-2019-neural}, who report an F1 worse than random guessing (macro F1 of 66.5 in ours, 17 in theirs). We believe this underperformance of finetuning BERT could be caused by a mistake in their training process. 

In Figure~\ref{fig:varyFewshotSizeResults}, we show the trend of performance on Reddit data as the number of samples increases. Prefix domain adaptation is comparable to finetuning, consistently outperforming regular prefix tuning. As shown by the larger shaded area around the lines, the stability of finetuning is worse than prefix domain adaptation for this task. Performance gradually converges increases as more data is given to each method.

\begin{table}
\centering
\resizebox{\linewidth}{!}{\begin{tabular}{c|cccc}
\hline
Fewshot Size & \textit{32} & \textit{64} & \textit{128} &\textit{256} \Tstrut \Bstrut \\
\hline
FT & 42.1\textsubscript{5.7} & 55.5\textsubscript{5.4} & 62.0\textsubscript{3.5} & \textbf{77.6\textsubscript{1.0}} \Tstrut \\
Domain Adapt + FT & 34.2\textsubscript{7.3} & 61.7\textsubscript{5.6} & \underline{66.6\textsubscript{8.7}} & \underline{77.3\textsubscript{1.3}} \Bstrut \\
\hline
Prefix Domain Adapt & \textbf{46.7\textsubscript{2.1}} & \textbf{63.5\textsubscript{1.5}} & \textbf{67.0\textsubscript{1.7}} & 72.2\textsubscript{1.0} \Tstrut \\ 
Prefix Adapt & 46.5\textsubscript{3.4} & \underline{63.1\textsubscript{2.5}} & 64.3\textsubscript{1.8} & 70.0\textsubscript{1.9} \\ 
P-Tuning v2 & \underline{46.7\textsubscript{1.7}} & 59.0\textsubscript{1.5} & 65.6\textsubscript{2.7} & 69.2\textsubscript{1.7} \Bstrut \\ 
\hline
\end{tabular}}
\caption{Classification results on RoBERTa-large, evaluated on Reddit data. Note that we do not evaluate results with LEGAL-BERT because LEGAL-BERT models with comparable size to RoBERTA-large do not exist.}
\label{tab:classification-large}
\end{table}
 
Larger models typically provide better performance on various tasks. Thus, we run experiments using RoBERTa-large (over 2x larger than RoBERTa-base) to see how our approach scales to larger models. As seen in Table~\ref{tab:classification-large}, our approach is still comparable to or outperforms full finetuning with larger models. Impressively, in the fewshot sizes 32-128, prefix domain adaptation with RoBERTa-base is even comparable to full finetuning with RoBERTa-large. Additionally, we note that full domain adapation is more sensitive to learning rates in larger models, explaining weaker performance in fewshot sizes 32 and 64. Due to limitations in computational resources, we leave more extensive hyperparameter search as future work.

\subsection{Calibration}
\begin{table}
\centering
\resizebox{\linewidth}{!}{\begin{tabular}{c|c|c|c}
\hline
& Reddit & LSE & ECHR \Tstrut\Bstrut\\
\hline
FT & 0.158\textsubscript{0.012} & 0.243\textsubscript{0.015} & 0.320\textsubscript{0.037} \Tstrut\\ 
LEGAL-BERT + FT & 0.454\textsubscript{0.05} & \textbf{0.165\textsubscript{0.043}} & \underline{0.245\textsubscript{0.042}} \\
Domain Adapt + FT & 0.152\textsubscript{0.004} & \underline{0.214\textsubscript{0.01}} & 0.320\textsubscript{0.121} \Bstrut\\ 
\hline
Prefix Domain Adapt & \underline{0.133\textsubscript{0.01}} & 0.242\textsubscript{0.023} & \textbf{0.214\textsubscript{0.032}} \Tstrut\\
Prefix Adapt & \textbf{0.104\textsubscript{0.021}} & 0.24\textsubscript{0.008} & 0.266\textsubscript{0.063} \\
P-Tuning v2 & 0.412\textsubscript{0.019} & 0.263\textsubscript{0.009} & 0.253\textsubscript{0.050} \Bstrut \\
\hline
\end{tabular}}

\caption{Calibration, measured by the top-1 expected calibration error (ECE). "Reddit" is the ECE on our Legal Advice Reddit dataset (fewshot size of 256), "ECHR" is ECE on European Court of Human Rights dataset (fewshot size of 32), and "LSE" is ECE on the Law Stack Exchange dataset (fewshot size of 256). Lower is better, with \textbf{bold} being the best and \underline{underline} being second best.}
\label{tab:calibration}
\end{table}

While providing predictions to laypersons, it is vital that the distribution of the output logits accurately reflect the model's confidence. Thus, we use the expected calibration error (ECE) \citep{naeini-etal-2015-obtaining} to measure the calibration of each model resulting from each method. We show that the calibration of our approach is better than finetuning across tasks, as seen in Table~\ref{tab:calibration}. Additionally, we observe that our approach is comparable to LEGAL-BERT across tasks. In the case where questions are well formulated (i.e., in the LSE dataset), we found that legal models are better calibrated. However, in Reddit data, which is central to helping laypersons with legal questions, we find that our approach is very competitive.

\subsection{Sample Efficiency}

\begin{figure}
    \centering
    \includegraphics[width=\linewidth]{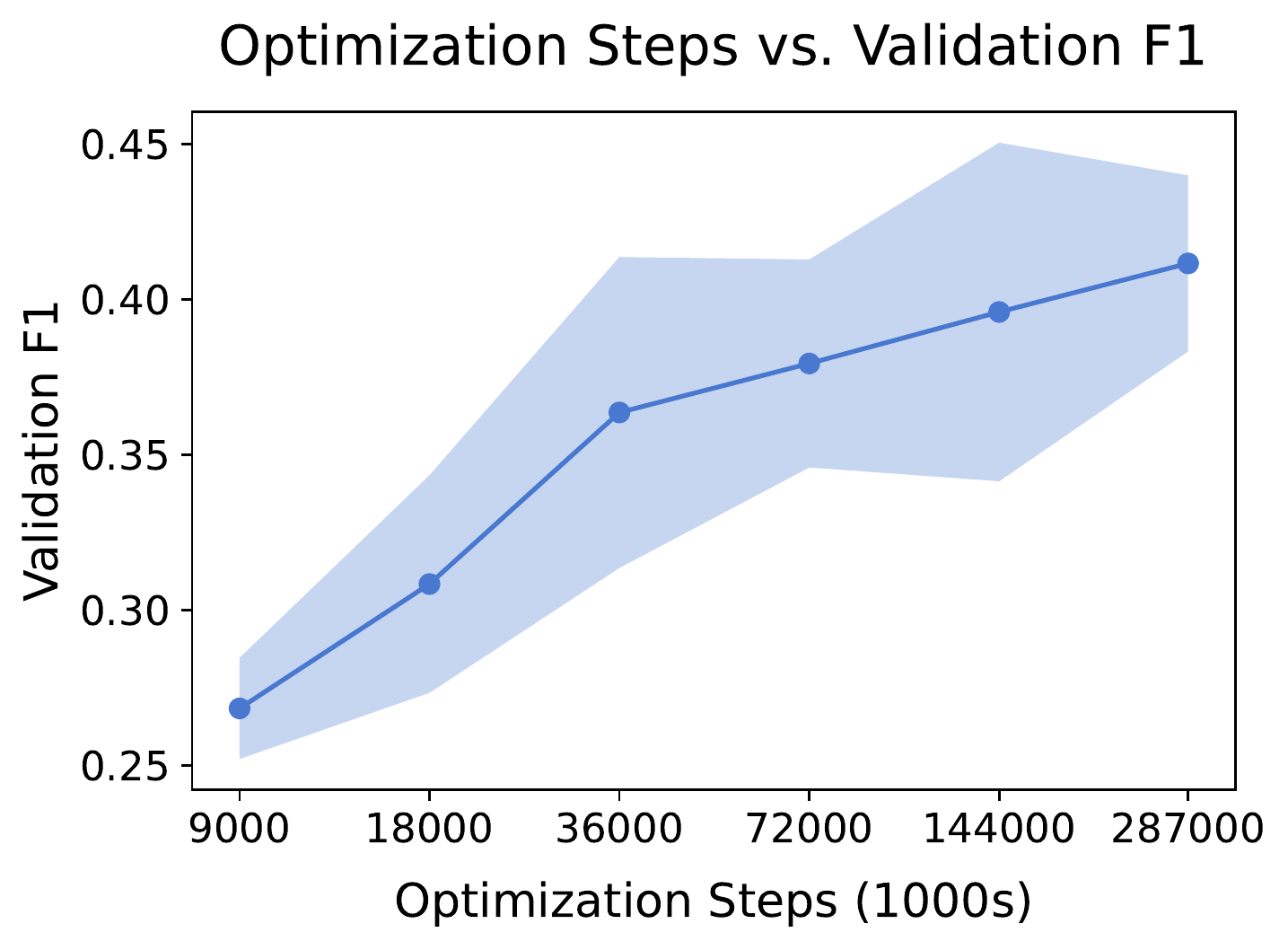}
    \caption{Performance of prefix domain adaptation after training the domain adapted prompt for a different number of training steps, performed on Reddit data with a fewshot size of 32 using RoBERTa-base. Shaded region represents standard deviation between five runs.}
    \label{fig:varyTrainingSteps}
\end{figure}

\begin{figure}
    \centering
    \includegraphics[width=\linewidth]{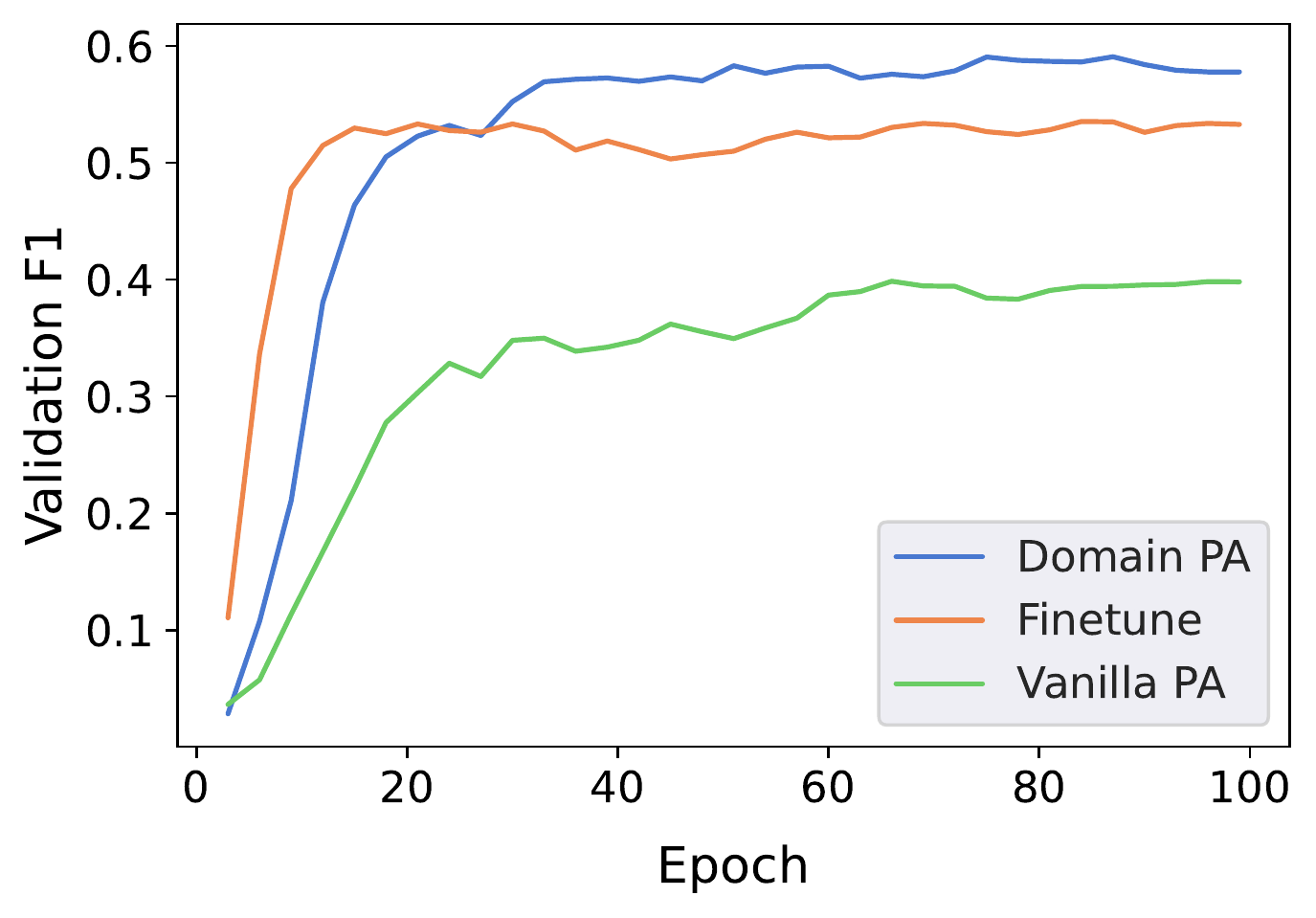}
    \caption{Convergence comparison of prefix domain adaptation ("Domain PA"), full finetuning ("Finetune"), and P-Tuning v2 on Reddit data, using a fewshot size of 64.}
    \label{fig:convergence}
\end{figure}

We study the effect of training time (i.e., number of training steps) for the domain-adapted prompt on downstream performance. To analyze the effect of additional training steps on the domain adapted prefix's performance, we initialize models using pre-trained prefixes from specific steps and plot the performance (over five runs) in Figure~\ref{fig:varyTrainingSteps}. We find that more optimization steps during the prefix adaptation step lead to better downstream performance. Intuitively, this makes sense as a longer training time means the prefix starts closer to an ideal one for a downstream task.

Though each optimization step is faster with regular prefix tuning \citep{gu-etal-2022-ppt}, it converges slowly and thus is not necessarily faster than finetuning. As shown in Figure~\ref{fig:convergence}, our approach converges faster than regular prefix tuning. Again, we argue that this is expected as the prompts are closer to a desired solution when compared to regular prefix tuning, meaning fewer training steps are needed to reach an effective solution.

\section{Conclusions}

In this paper, we propose a novel training framework, \textit{prefix domain adaptation}, aiming to domain adapt a prompt using a large corpus of domain-specific text. We show that our approach matches or outperforms LEGAL-BERT or related techniques in performance while training fewer (0.1\%) parameters. With our technique, we improve fewshot performance and convergence time compared to other parameter-efficient methods. We believe this will make fewshot data more usable (and thus reduce data labelling costs) while using parameter-efficient methods to reduce computational and storage costs.

Additionally, we introduce two new datasets (Legal Advice Reddit and Law Stack Exchange) to lay foundations for future work in legal decision-making systems; as opposed to formal documents in ECHR, our two datasets are closer to legal questions asked by laypersons, helping to promote access to justice for all.

\bibliography{anthology,custom}

\appendix

\section{Additional Training Details}
\label{sec:trainingDetails}

\begin{table}[t]
    \begin{tabular}{c|c}
        \hline
        Configuration & Learning Rates \Tstrut \Bstrut \\
        \hline
        RoBERTa-base PT &  5e-2, 3e-2, 2e-2, 5e-3, 5e-4 \\
        RoBERTa-large PT & 5e-2, 3e-2, 2e-2, 5e-3, 5e-4 \\
        RoBERTa-base FT &  1e-3, 5e-4, 2e-4, 1e-4, 5e-5 \\
        RoBERTa-large FT & 1e-4, 5e-5, 2e-5, 1e-5, 5e-6 \Bstrut \\
        \hline
    \end{tabular}
    \caption{Learning rates searched for each configuration. The suffix "PT" means for prompt tuning based methods, and "FT" for finetuning based methods.}
    \label{tab:learningRates}
\end{table}

We use the AdamW optimizer and a grid search of learning rates as in Table \ref{tab:learningRates}, mostly following \citet{gu-etal-2022-ppt}. For all of our experiments, we truncate the sequence to a length of 500 tokens (as opposed to 512 tokens) to allow space for a tuned deep prefix prompt. We report the calibration and general results using the checkpoint with the best validation macro F1, for each fewshot size and method.

Given that RoBERTa-base (\textasciitilde125M parameters) and RoBERTa-large (\textasciitilde355M parameters) can fit in a single NVIDIA 1080Ti GPU (using a smaller batch size), we do not perform any model or data parallelism. We use an effective batch size (i.e., factoring in gradient accumulation steps) of 32 for experiments on roberta-base, and due to memory constraints, an effective batch size of 24 for experiments on roberta-large. As the number of samples is low, we train for 100 epochs. However, while performing domain adaptation and prefix adaptation training steps, we train for 20 epochs as much more data as available (and therefore, more optimization steps are run in each epoch).

We use a prefix length of 8. Including the tuned linear head for classification, the largest number of parameters we tune for RoBERTa-base is 160K (varies slightly for each task depending on the number of classes), or \textasciitilde0.13\% of the model's parameters.

\section{Data Details}
\label{sec:dataDetails}

\begin{table}
\centering
\resizebox{\linewidth}{!}{
\begin{tabular}{c|c|c|c|c}
\hline
Dataset Name & $N_{train}$ & $N_{dev}$ & $N_{test}$ & Avg. Words \Tstrut\Bstrut\\
\hline
ECHR & 7100 & 2998 & 1380 & $2105_{2489}$ \Tstrut\\
Legal Advice Reddit & 9887 & 9987 & 79136 & $145_{117}$ \\
Law Stack Exchange & 638 & 319 & 1596 & $244_{217}$ \Bstrut \\
\hline
\end{tabular}}
\caption{Sizes of datasets. $N_{train,dev,test}$ represent sizes of the train, development, and test sets respectively.}
\label{tab:dataSize}
\end{table}

For Reddit data we take the top 11 classes that are not countries. We concatenate the title of the Reddit post and body text together, then use this combination to train our models for the masked language modelling and flair classification task.

For Stack Exchange data, we take only the questions with a single tag, and again. The stack exchange data, taken from Internet Archive\footnote{\href{https://archive.org/download/stackexchange}{https://archive.org/download/stackexchange}}, includes the post body in an HTML form. As our base models were not trained on HTML formatted text, we convert the HTML to Markdown footnote to make it much more similar to human readable text.

For the ECHR dataset, we use the non-anonymized variant and concatenate the title of the case with each fact from the legal case. Additionally, we found that some documents had numbered facts (such as "\textbf{1.} <fact>"), while some documents were not numbered. We used a simple regular expression to remove this inconsistency which could possibly create biases in the model (e.g., if numbered facts were more likely to mean a violation).

In our domain adaptation experiments, we use all the data (i.e., including questions/posts that were previously filtered out because they didn't have top tags) for each dataset. We use the domain adapated checkpoint with the best validation cross-entropy loss for downstream tasks.

The sizes of each split are listed in Table~\ref{tab:dataSize}. Test split sizes for the Reddit and Stack Exchange dataset are intentionally larger than the validation and training set to better simulate true fewshot learning, as per \citet{perez-etal-2021-true}.

\end{document}